\documentclass[journal,comsoc]{IEEEtran}
\usepackage[T1]{fontenc}

\usepackage{cite}

\ifCLASSINFOpdf
  \usepackage[pdftex]{graphicx}
  \graphicspath{{../pdf/}{../jpeg/}}
  \DeclareGraphicsExtensions{.pdf,.jpeg,.png}
\else
   \usepackage[dvips]{graphicx}
   \graphicspath{{../eps/}}
   \DeclareGraphicsExtensions{.eps}
\fi
\usepackage{color}

\usepackage{amsmath}
\usepackage[cmintegrals]{newtxmath}
\usepackage{array}
\usepackage{multirow}
\usepackage{threeparttable}
\begin{document}
%
\title{An Artificial Intelligence-Based System to Assess Nutrient Intake for Hospitalised Patients}
%
%
%

\author{Ya~Lu,
        Thomai~Stathopoulou,
        Maria F.~Vasiloglou,
        Stergios~Christodoulidis,
        Zeno~Stanga,
        and~Stavroula~Mougiakakou
\thanks{Research supported by SV Foundation.}
\thanks{Ya Lu, Thomai Stathopoulou, Maria F. Vasiloglou, Stergios Christodoulidis and Stavroula G. Mougiakakou are with the ARTORG Center for Biomedical Engineering Research, University of Bern. Murtenstrasse 50, CH-3008 Bern, Switzerland (phone: +41-31-632-7592; e-mail: 
({ya.lu, thomai.stathopoulou, maria.vasiloglou, stergios.christodoulidis, stavroula.mougiakakou}@artorg.unibe.ch).
}
\thanks{Zeno Stanga is with the University Hospital of Diabetology, Endocrinology, Nutritional Medicine and Metabolism (UDEM) Inselspital, University Hospital, Freiburgstrasse 15, CH-3010 Bern, Switzerland (e-mail: zeno.stanga@insel.ch).}}

%
%

\markboth{Journal of \LaTeX\ Class Files,~Vol.~14, No.~8, August~2015}%
{Shell \MakeLowercase{\textit{et al.}}: Bare Demo of IEEEtran.cls for IEEE Communications Society Journals}
%



\maketitle

\begin{abstract}
Regular monitoring of nutrient intake in hospitalised patients plays a critical role in reducing the risk of disease-related malnutrition. Although several methods to estimate nutrient intake have been developed, there is still a clear demand for a more reliable and fully automated technique, as this could improve data accuracy and reduce both the burden on participants and health costs. In this paper, we propose a novel system based on artificial intelligence (AI) to accurately estimate nutrient intake, by simply processing RGB Depth (RGB-D) image pairs captured before and after meal consumption. The system includes a novel multi-task contextual network for food segmentation, a few-shot learning-based classifier built by limited training samples for food recognition, and an algorithm for 3D surface construction. This allows sequential food segmentation, recognition, and estimation of the consumed food volume, permitting fully automatic estimation of the nutrient intake for each meal. For the development and evaluation of the system, a dedicated new database containing images and nutrient recipes of 322 meals is assembled, coupled to data annotation using innovative strategies. Experimental results demonstrate that the estimated nutrient intake is highly correlated (> 0.91) to the ground truth and shows very small mean relative errors (< 20\%), outperforming existing techniques proposed for nutrient intake assessment.
\end{abstract}

\begin{IEEEkeywords}
Artificial Intelligence, nutrient intake assessment, few-shot learning.
\end{IEEEkeywords}

%
\IEEEpeerreviewmaketitle

\section{Introduction}
%
%
%
%
\begin{figure*}[htb]
\centering
\includegraphics[width=1.0\linewidth]{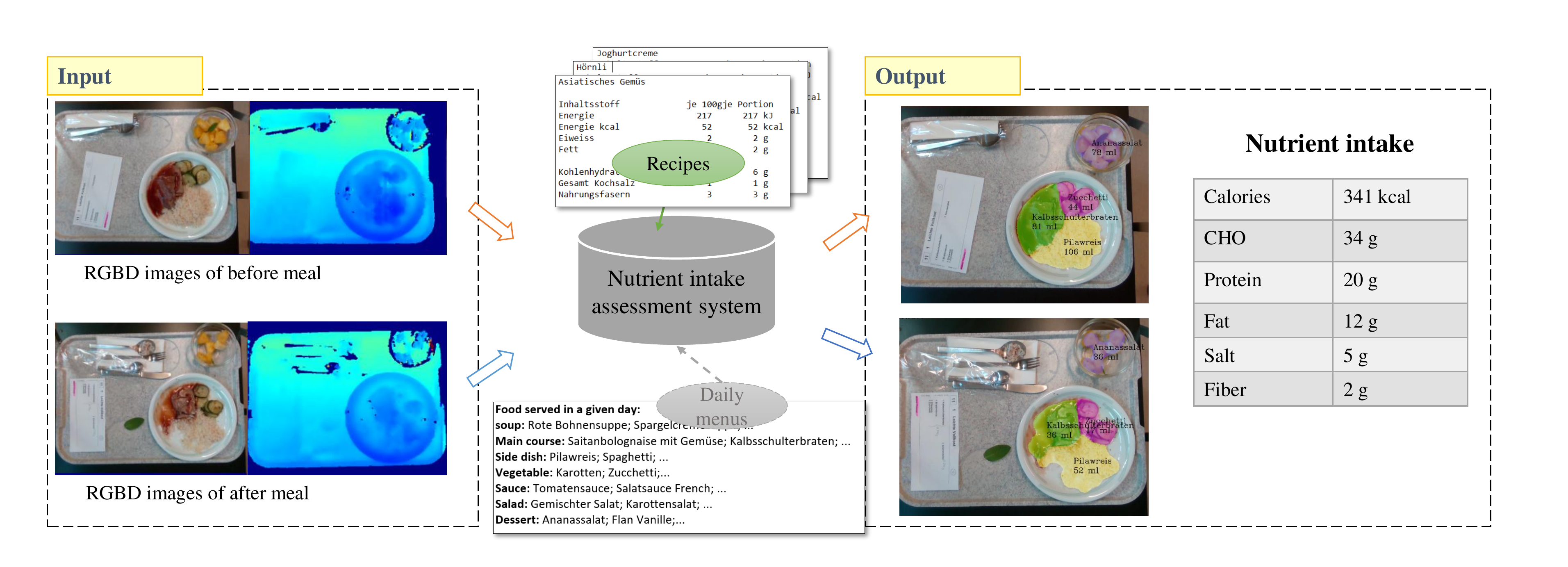}
   \caption{An outline of the proposed system for assessing nutrient. The inputs of the system are the RGB-D image pairs captured before and after the meal. The recipes include nutrient information and may be  pre-integrated into the system. Daily menus can be integrated subsequently. The outputs of the system are the results of food segmentation, recognition and volume estimation before and after the meal. By analysing these results, the nutrient intake of the meal can be estimated.}
\label{fig:outline}
\end{figure*}

\IEEEPARstart{M}{ALNUTRITION} of hospitalised patients is a serious condition associated with an increased risk of hospital infections, higher mortality, morbidity, prolonged length of hospital stay and extra healthcare expenses \cite{study_1}. Investigations performed among hospitalised patients in different countries have shown that the average prevalence of hospitalised malnutrition may be in the order of 40\% \cite{study_2, study_3, study_4, study_5}. Thus, maintaining good nutritional status is of vital importance for both hospitalised patients and social medical systems. \par
Since hospitalised malnutrition is mainly attributed to the poor recognition and monitoring of nutritional intake \cite{study_6}, it is crucial to regularly evaluate the daily food intake of hospitalised patients. Traditionally, this relies on non-automated or semi-automated approaches, such as food weighing \cite{study_6, study_7}, visual estimations \cite{study_7, study_8, study_9}, or digital photography \cite{study_6}, \cite{study_10}, which are either time consuming, expensive or prone to errors. Therefore, there is still an unambiguous need for a reliable and simple solution to assess nutrient intake.\par
With the growth in AI-based image processing methods, it has become feasible to analyse food items through a food image, so that there is great potential to make estimation of nutrient intake fully automatic. Recently, AI-based dietary assessment systems, such as Im2Calories for calorie estimation \cite{im2calories} and GoCARB for carbohydrate (CHO) estimation \cite{gocarb}, have been proposed. The systems process food images in three steps: 1) food items segmentation; 2) food recognition and 3) food volume estimation. Thus, nutrient content can be calculated from the food nutrient database. While these pioneer studies have demonstrated their viability to evaluate the food intake using AI, there is also a need to further improve the performance in terms of the accuracy of estimation and the number of supported food categories. However, the difficulty lies in the fact that the sophisticated annotation requirements intrinsically limit the quality and the size of the food images database for nutrient intake assessment. This further impedes the use of some advanced AI algorithms \cite{mariosfoodclass,wide-slice-class,Aguilar_recog} that have already been applied for food image analysis, but which heavily depend on having a large database. Therefore, we require a dedicated database and associated AI algorithms that can be adapted to limited training data. \par
In this paper, which is a continuation of our work in \cite{embc2019}, we propose an AI-based, fully automatic system for assessing nutrient intake for hospitalised patients, by processing RGB-D image pairs captured before and after a meal. A new database for food images with associated nutrient information and recipes is created from 322 food trays (meals) as prepared for hospitalised patients by the central kitchen of Bern University Hospital. The database includes a total of 1281 food items from 521 food categories. This database is used for the design, development and validation of a number of AI-based algorithms to estimate nutrient intake. It involves: 1) a Multi-Task Contextual Network (MTCNet) for food segmentation, which employs a pyramid architecture for feature encoding and a newly designed CTLayer to provide contextual information between foods and serving plates; 2) a novel, in terms of architecture, few-shot learning-based classifier \cite{prototypicalnet} for food recognition. The food recognition model is trained within the framework of meta-learning \cite{prototypicalnet} and takes advantage of the transferred weight; 3) a 3D surface construction algorithm \cite{joachim3d} for the estimation of the consumed food volume; and 4) a nutrient intake calculator that links the consumed volume, food type and recipe. Experimental results demonstrate very strong correlation coefficients between the software prediction and the ground truth ($r>0.91$ for calories and all nutrient types; $p<0.001$), and very small mean relative errors ($< 20$\% for calories and all the nutrient types). To the best of our knowledge, this outperforms previous studies. \par
Fig. 1 presents an outline of the proposed system. The inputs of the system are the RGB-D image pairs captured before and after the meal. The outputs of the system are the estimated food segmentation, recognition and volume, for the cases of before and after the meal, along with the nutrient intake estimations.\footnote{The food category name of the system is in German.} \par
Our contributions are summarised as follows: 

\begin{enumerate}
\item We propose the first AI-based, fully automatic pipeline system to assess nutrient intake for hospitalised patients.
\item We have built a new database that contains food images with associated recipes and nutrient information. The database has been collected in real hospital scenarios.
\item We have designed and developed an innovative series of AI algorithms that can provide good performance on small quantity training data. The newly developed algorithms include, for instance, the Multi-Task Contextual Network (MTCNet) for food segmentation, and few-shot learning based classifier for food recognition. Extensive experiments have been conducted to demonstrate the advantages of these proposed methods.
\end{enumerate}
\section{Related work}
\subsection{Assessment of nutrient intake in hospitalised patients}
Weighing the food before and after each meal is currently the most accurate method to monitor nutrient intake\cite{study_6}. However, this is not adaptable to large samples or multicentre studies, due to the complexity of its implementation . As it is highly accurate, this method has increasingly often been taken as a reference method in studies assessing nutrients \cite{study_6,study_7,study_8, study_9,study_10}. \par
Visual estimation is a method that approximately estimates the percentage of the food intake by visually observing food left and comparing this with a reference scale. This eliminates the manual operation in the food weighing approach. For hospitalised patients, visual estimation scales that have been used include a 4-point scale (all, $1/2$, $1/4$, none) \cite{study_8} and a 3-point scale (all, k$ >50\%$, $<50\%$) \cite{study_9}. Although such a method has been widely used because of its simplicity \cite{study_19}, studies have indicated that the estimation of food intake is often inaccurate, which commonly results in overestimation of $15\%$ \cite{study_19}. \par
Instead of observing the test meal, more recent approaches - known as digital photography - estimate food intake on the basis of the food images captured before and after the meals, using either cameras \cite{study_6} or smartphones \cite{study_10}. The visual assessment can then be performed remotely by professional healthcare staff under free-living conditions. Despite the clear progress with respect to visual estimation, methods based on digital photography are still either semi-automated or non-automated, and are subject to human estimation. \par
It is worth mentioning that all these approaches assume that all the involved food types are correctly recognised, which is however too optimistic in real applications.
\subsection{Food Database}
Currently, most of the food databases are built for the food recognition task, and usually annotate the food images with image-level labels. The first published food recognition database is the Pittsburgh Fast Food Image Dataset (PFID) \cite{PFID_food}, which contains 101 fast food categories. After this, the larger Food101 \cite{food101} database is presented, and contains 101,000 images belonging to 101 food classes. A large-scale food image database \cite{ifood_data}, involving 158,000 images from 251 fine-grained food categories, has recently been proposed and used for the ``IFOOD2019'' food classification challenge \cite{ifood_challenge}. Furthermore, because meals are characterized by occlusions of ingredients, and high intra-class and low inter-class variability between food classes, the recipes that include food ingredients can be used for better understanding of the food categories. So far, the largest image-recipe database is Recipe1M+ \cite{recipe1m}, which contains more than 1 million cooking recipes and 13 million food images in total.\par
Databases for food detection/segmentation are distinct from food recognition databases and require additional annotation of food location on the images. UEC Food 256 \cite{uec256}, which includes 256 Asian food categories with 31,651 images, provides both image-level labels and the corresponding food bounding box for each image. On the other hand, the UNIMIB2016 database \cite{unim_data} includes 1027 food images with 73 food categories and annotates the food locations with pixel-level food segmentation map. \par
However, there are very few databases built for food volume estimation. The annotations required in such databases are image level labels, the pixel-level food segmentation map, along with the food volume ground truth. One such database is the MADiMa2017 database \cite{madima17}, which involves 234 food items from 80 central European meals. The image labels and the food locations are annotated using a pixel-level food semantic segmentation map, while the volume ground truth of each food item is built using online AutoCAD.\par
For nutrient calculation, the USDA nutrient database \cite{usda} is the one most commonly used to translate the food type and volume to nutrient content. However it may not include all the target food types of an AI-based system and cannot easily be used outside of the US.\par
To summarise, the complexity of the sophisticated annotation requirements intrinsically limits the quality and the size of the food images database for assessment of nutrient intake. This further impedes the use of some advanced AI algorithms that rely on a large amount of data. Thus, it is clearly desirable to develop dedicated algorithms adapted to the limited available databases.

\subsection{AI-based dietary image processing}
Traditional food segmentation methods are based on region growing and merging algorithms, along with handcrafted features, e.g. \cite{marios_seg, early_Wearable}, which are subsequently improved to employ a border detection network \cite{joachim_seg}. Although such methods are fast and easy to implement, they are limited in some complex scenarios, such as noisy background, similar food items and imperfect light conditions. More recently, a fully convolutional network (FCN) has been applied for food-nonfood segmentation \cite{grabpay}, which performs better than traditional methods. However, it is not adapted to applications requiring food-food segmentation.\par
Food recognition at the early stage relies on the handcraft features (e.g. colour, texture and SIFT) \cite{mariosfoodclass, early_Wearable} and traditional classifiers, such as SVM and ANN \cite{mariosfoodclass, early_Wearable}. The performance of the food recognition has recently been significantly improved by CNN-based approaches \cite{wide-slice-class, Aguilar_recog, ifood_challenge}, which have demonstrated an accuracy of 90.27\% on the Food101 database \cite{wide-slice-class} and 94.4\% top3 accuracy in the IFOOD2019 challenge \cite{ifood_challenge}. However, these existing CNN-based methods require a large amount of training data to achieve high accuracy, which is costly and impractical for a given application that may only provide limited data size. Although a nearest neighbour-based method that uses few training samples is being investigated for personalised food recognition \cite{personalised}, its performance still largely depends on the original base classifier that requires a large amount of training data.\par
Most of the approaches for estimating food volume are based on geometry \cite{joachim3d,early_volume}, and require multiple RGB images and a reference object as input for 3D food model construction, and these are not robust to low texture food and large view changes. Food volume can also be estimated using the CNN-based regression method \cite{madima2018}, while the performance depends heavily on the quantity of the training data. With the development of high-quality depth sensors or stereo cameras on smartphones, depth maps can be utilised for estimating food volume with no need for reference object or extensive training data \cite{madima17, yanaicalories, volume_synthesis}; this provides more stable and more accurate results than with geometry-based approaches \cite{joachim3d,madima17,early_volume}.\par
\begin{figure}[t]
\centering
\includegraphics[width=1.0\linewidth]{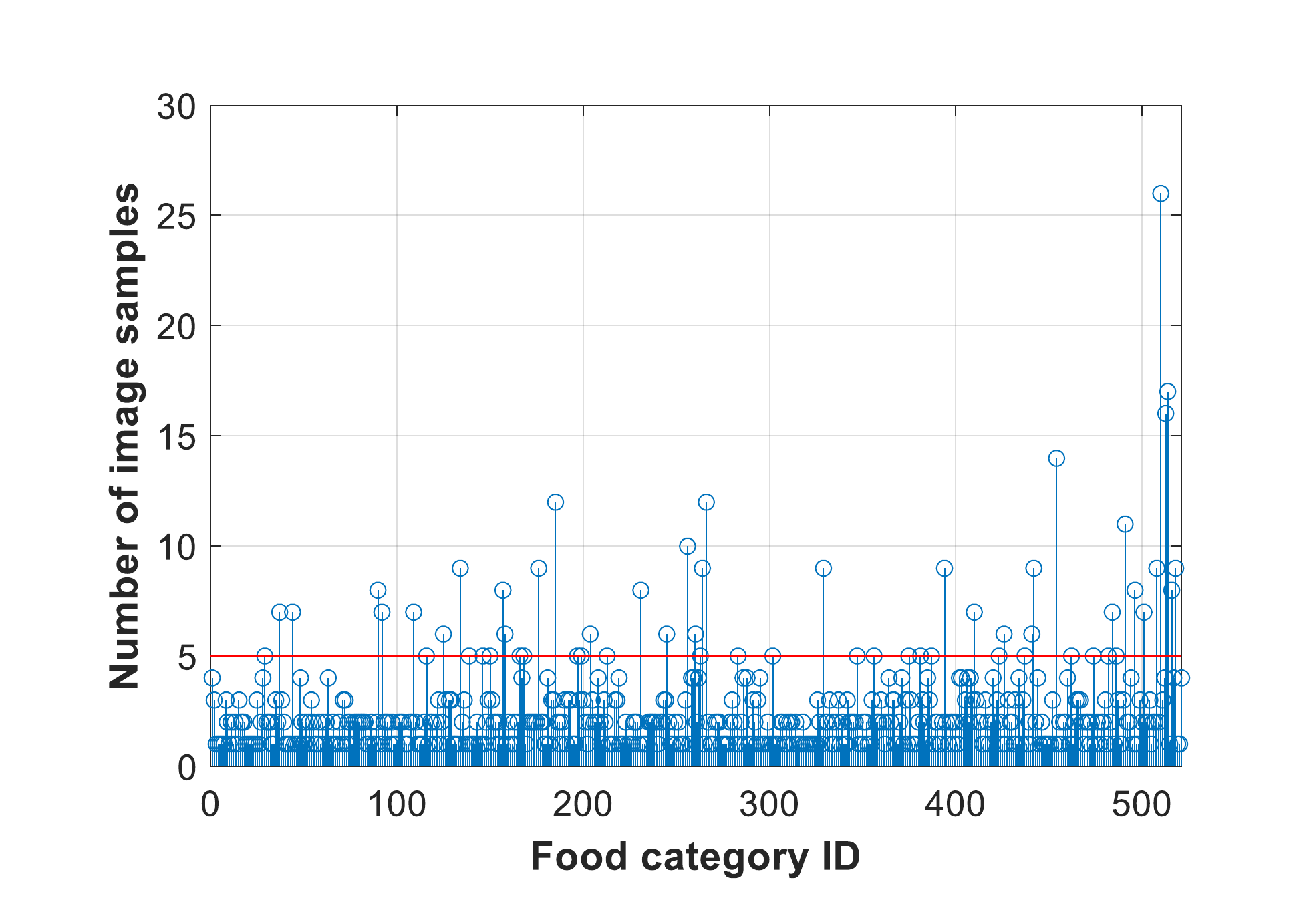}
   \caption{Statistics of the number of images contained in one category. 521 fine-grained food categories are included in the database. There are fewer than 5 image samples for most categories.}
\label{fig:dbs}
\end{figure}
\begin{figure}[t]
\centering
\includegraphics[width=1.0\linewidth]{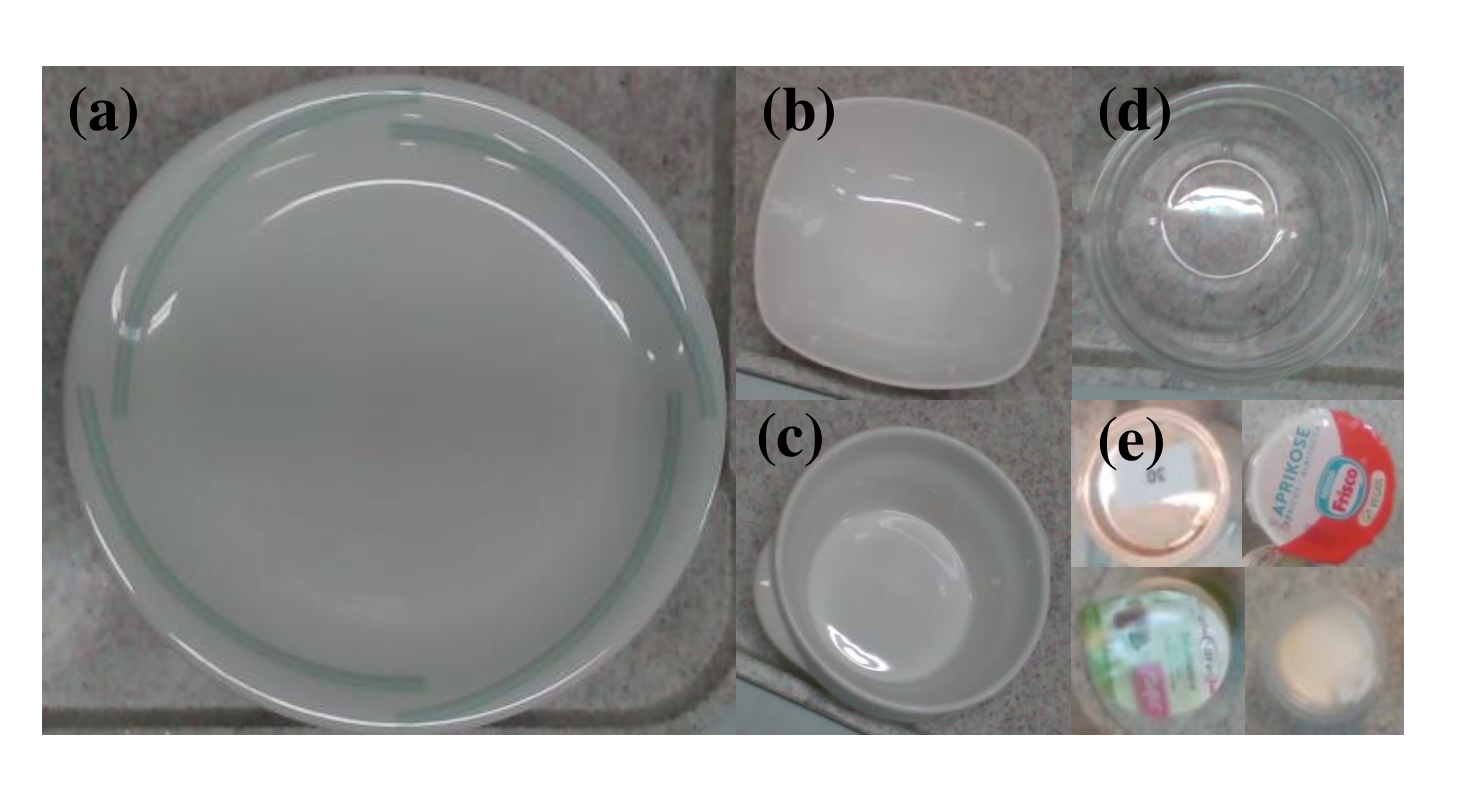}
   \caption{Plates used in Bern Hospital Kitchen; (a): plates for main course, vegetable, side dish; (b): salad bowl; (c): soup bowl; (d): dessert bowl; (e): other packaged containers.}
\label{fig:plate}
\end{figure}
\section{Data set}
\begin{figure*}[t]
\centering
\includegraphics[width=7in]{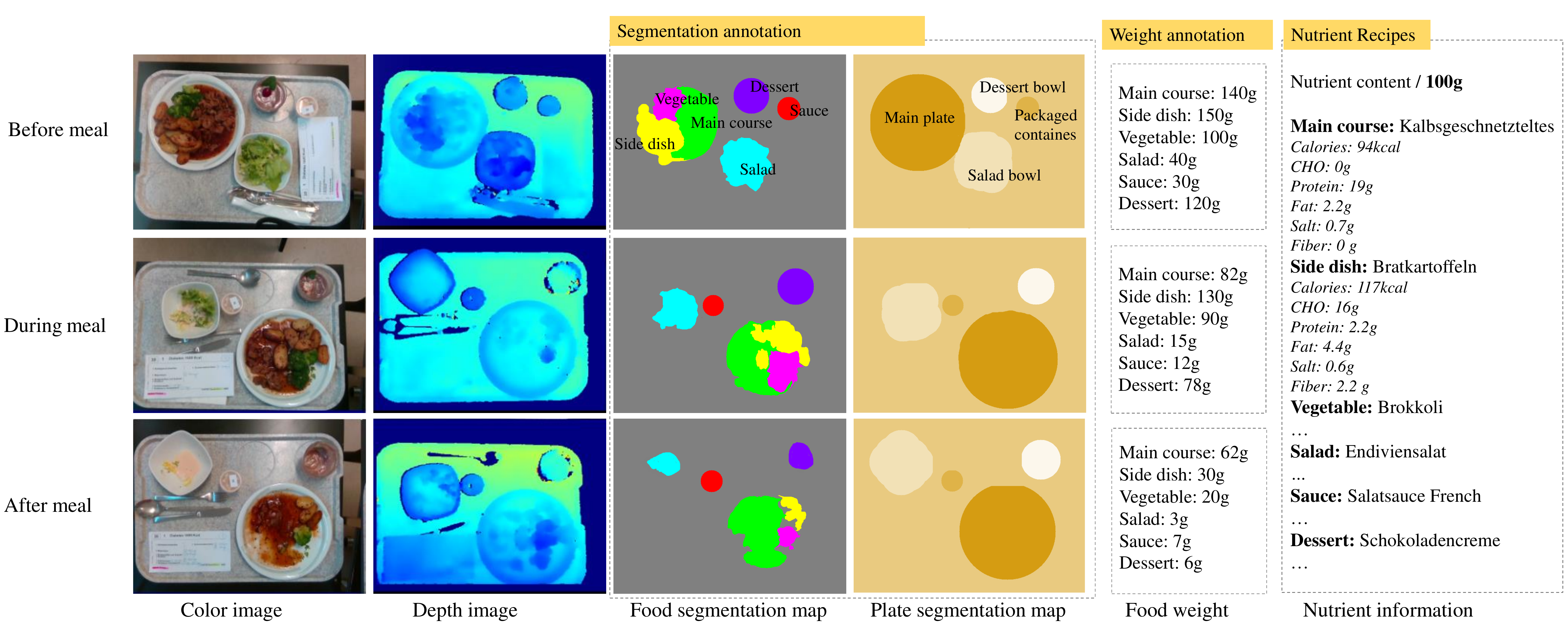}
   \caption{One typical meal in the NIAD database. The food items are segmented into 7 hyper categories, while the plates are classified into 5 types during the food image annotation. The nutrient ground truth is annotated according to the recorded weight and nutrient information provided by the hospital kitchen. }
\label{fig:dbd}
\end{figure*}
For the design, development and evaluation of various components of the proposed system, a dedicated and novel database, named ``Nutrient Intake Assessment Database'' (NIAD), was compiled, which contains RGB-D image pairs of 322 real world meals, including 1281 food items associated to 521 food categories in total. All the meals -as well as the associated portion size, nutrient information, recipes and menus-  were provided by the central kitchen of  Bern University Hospital. During a 2-month procedure for data collection, images of 1\(\sim\)15 meals involving 3\(\sim\)36 food categories were captured each day.
\subsection{Data capture}
The image data was captured using the Intel RealSense RGB-D sensor, which outputs aligned RGB and depth images simultaneously. The distance from the camera to the food tray was randomly selected between 35cm and 50cm during image capture. The resolution of each acquired RGB-D image pair is $480\times640$ (Height\(\times\)Width).\par
To ensure that the database is highly diverse, the acquisition of the image pairs was conducted in two stages. Firstly, 153 RGB-D image pairs were captured from 153 food trays only before each meal; then another 507 RGB-D image pairs were captured from the remaining 169 food trays, before, during and after each meal. Note that the purpose of capturing the image pairs during the meal is to mimic the possible scenario of patients with little appetite in the real hospital. While capturing all these 660 RGB-D image pairs, the weight of each plate inside the food tray was recorded using scales with the unit of gram. This will hereafter be used to annotate the nutrient intake ground truth.
\subsection{Data Annotation}
\begin{figure*}[t]
\centering
\includegraphics[width=1.0\linewidth]{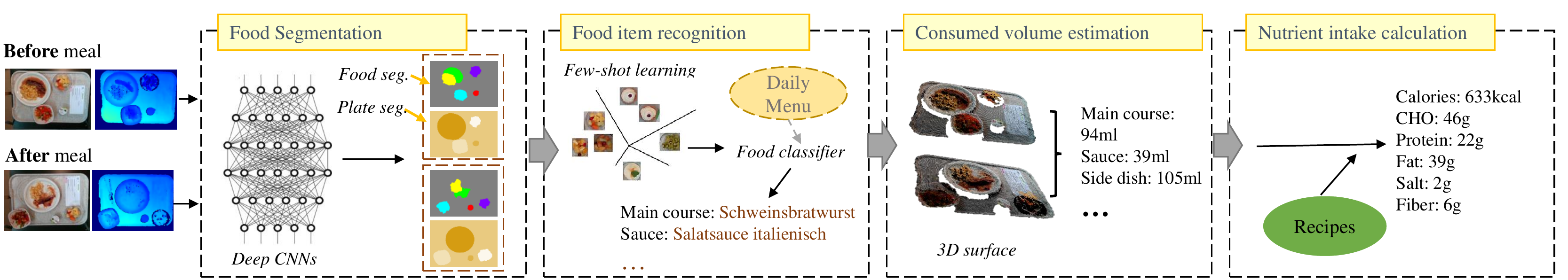}
   \caption{Overview of the proposed system. Four main steps are sequentially performed on the RGB-D image pairs captured before and after the meal to assess the nutrient intake: food segmentation, food item recognition, consumed volume estimation and nutrient intake calculation.}
\label{fig:brief}
\end{figure*}
To create the ground truth for the food image analysis and the assessment of nutrient intake, we annotated the captured data in both image level and nutrient level.
\subsubsection{Food image annotation}
The pixel-level ground truth semantic segmentation map of each food tray image was annotated using an in-house developed image annotation tool (Appendix \ref{app:tool} contains information regarding the tool). 
According to the menus provided by the hospital kitchen, the segmented food items were annotated into 7 hyper categories (main course, side dish, vegetable, sauce, soup, salad and dessert), corresponding to 521 fine-grained categories (belonging to 139 main courses, 86 side dishes, 42 vegetables, 46 soups, 32 salads, 96 sauces and 80 desserts), which are further used for the refined recognition. It is worth mentioning that, although there is a large number of food categories (521) in the dataset, most of these only correspond to few image samples ($<5$), as summarized in Fig. \ref{fig:dbs}.\par
In addition to food segmentation, we also dedicatedly segmented plates and classified the plate types into five (5) categories: main plate, salad bowl, soup bowl, dessert bowl, and other packaged containers, as exemplified in Fig. \ref{fig:plate}. It should be emphasised that the use of such plate segmentation maps can greatly improve the performance of the proposed system in three aspects: 1) allowing the use of a novel multi-task learning approach to improve the accuracy of food segmentation; 2) providing the contextual information to further refine the food semantic segmentation map in the post-processing stage; 3) enabling estimation of the depth of plate surface, in order to correctly calculate the food volume. All these benefits will be demonstrated in Section \ref{sec:exp}.
\subsubsection{Nutrient annotation}
For images captured before the meals, the nutrient content of each food item is annotated using the weight and nutrient information provided by the hospital kitchen. For the images captured during and after meals, two annotation strategies are applied - depending on how the food is served: 1) when one food item is served in one plate, the net weight of the food is calculated by subtracting the weight of the empty plate from the recorded plate weight. The nutrient is annotated according to the weight difference with respect to the empty plate and using the nutrient content per 100 grams given by the recipes; 2) when more than one food items are served in one plate, the iterated plate weight difference and the visual estimation approach \cite{plate_gt_g} are adopted for nutrient annotation. This procedure gives almost identical results to the weighted food method ($r=0.99; p<0.001$) and can be obtained even by untrained annotators.
Fig. \ref{fig:dbd} exemplifies one of the typical meals in the database, including the images captured before, during and after meals, together with the annotations.

\section{Methodology}
\label{sec:method}
The overall architecture of the proposed methodology is illustrated in Fig. \ref{fig:brief}. In the following sub-sections, the four main stages involved in the proposed system are described in detail.

\subsection{Food segmentation}
\label{sec:method:seg}
\begin{figure}[t]
\centering
\includegraphics[width=1.0\linewidth]{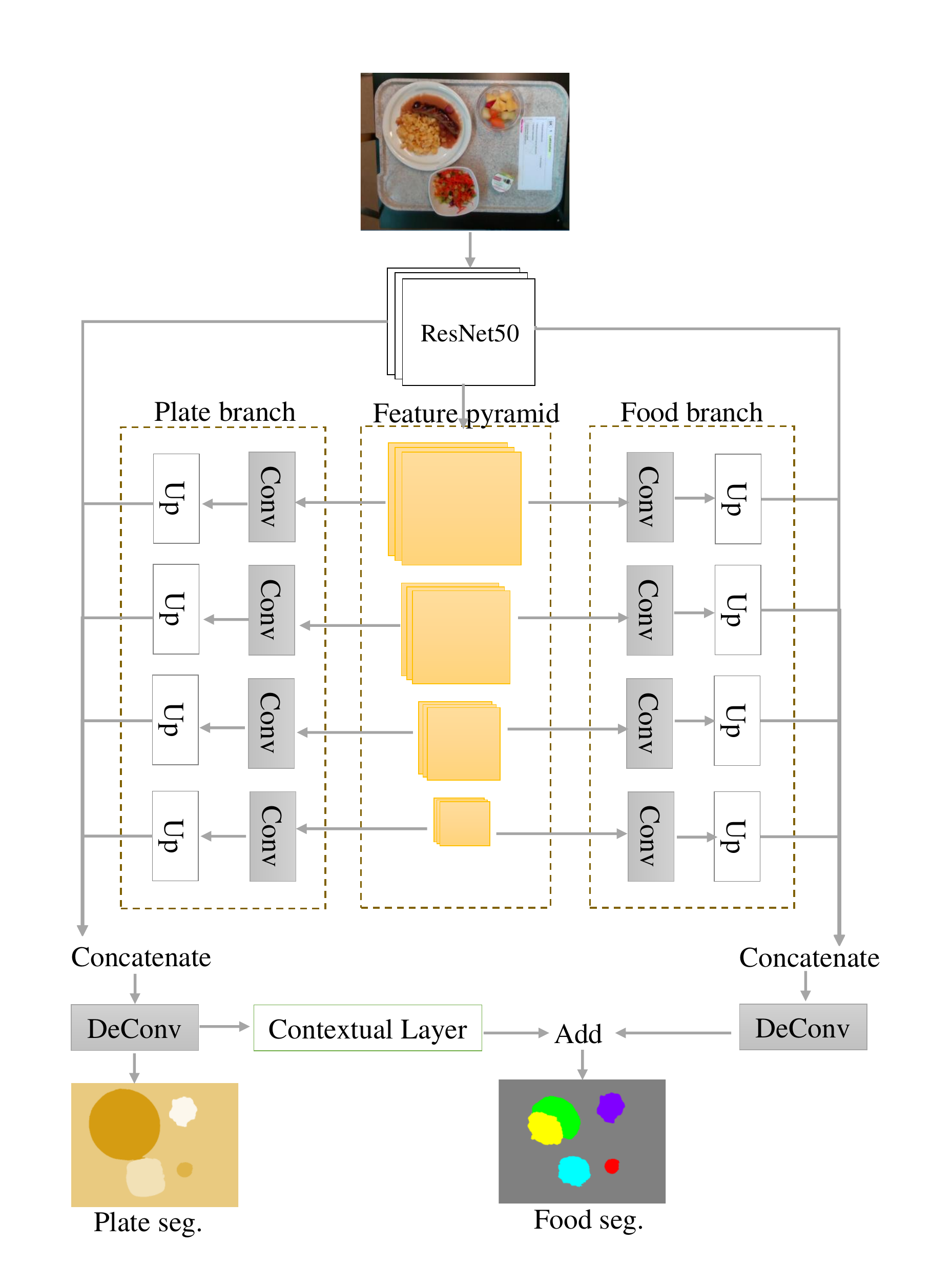}
   \caption{Architecture of food segmentation network. ``Conv'' and ``DeConv'' mean convolutional layer and deconvolutional layer, respectively, while ``Up'' indicates up-sampling layer }
\label{fig:arc}
\end{figure}
A Multi-Task Contextual Network (MTCNet) is newly proposed for image segmentation, which takes the colour image as input, and outputs the segmentation maps of the food type\footnote{All the ``food type'' in this section indicate ``hyper food type''.} and the plate type simultaneously, as described in Fig. \ref{fig:arc}. This network employs a pyramid feature map fusion architecture \cite{spp_he,pspnet} that features a large receptive field, and is thus able to overcome the problem of confusing the food type -as commonly occurs in semantic segmentation algorithms \cite{segnet,FCN}. In addition, the contextual relation between the food and plate type is enhanced by a newly proposed Contextual layer (CTLayer), which further improves segmentation accuracy. These two advantages are both experimentally demonstrated in Section \ref{sec:exp}. In the current section, we focus on elaborating the detailed configuration of the proposed network.\par
Firstly, an initial feature map with size of $60\times80\times2048$ (Height$\times$Width$\times$Channel) is generated using a pretrained dilated ResNet50 \cite{he_res,dilated}. By applying the ``average pooling'' on this feature map with 4 different pulling sizes, 4 pyramid feature maps are generated with sizes of $15\times20\times2048$, $8\times10\times2048$, $3\times4\times2048$, $1\times1\times2048$, respectively. \par
Then, for both ``food'' and ``plate'' segmentation branches, the pyramid feature maps of each level are fused and concatenated with the initial feature maps (from ResNet50) through a convolutional layer and an up-sampling layer (which resizes the feature map using interpolation). Like the strategy used in \cite{pspnet}, the convolutional layers we used here are all with $1\times1$ kernel size and 512 output channels (which equals to $1/4$ of the initial channels), in order to maintain the relative weight of the initially encoded features. Finally, two deconvolutional layers with the same kernel size of $8\times8$ are applied for both ``food'' and ``plate'' predictions, with channel numbers of 8 (food type+1 for background) and 6 (plate type+1), respectively, corresponding to 2 outputs with sizes of $480\times640\times8$ and $480\times640\times6$. Note that the final ``food'' prediction - indicating image segments of each food item and the corresponding hyper food category - has incorporated the contextual relation between the food and the plate provided by a CTLayer - which is basically a convolutional layer adopted upon the ``plate'' prediction, with $3\times3$ kernel size and 8 output channels. \par
All the above mentioned convolutional/deconvolutional layers are followed by batch normalisation layers and RELU activation layers except the two prediction layers, each of which is followed by a ``softmax'' layer.

\subsection{Food item recognition}
Using the approach introduced in section \ref{sec:method:seg}, the hyper food semantic segmentation has been predicted. In this section, the fine-grained food categories are recognised by further processing the predicted hyper food segments. Since the fact that the limited image samples of each fine-grained food category hinders the use of typical networks\cite{wide-slice-class,Aguilar_recog}, we have designed a novel few-shot learning-based classifier that requires only few annotated samples of each category.\par
\subsubsection{Model}
The few-shot learning-based classifier is trained within the framework of meta-learning \cite{prototypicalnet}. The key idea is to learn the transferred knowledge among a large number of similar few-shot tasks, which can be further used for the new tasks \cite{prototypicalnet,match_net,relation_net}. Each few-shot task includes a support set and a query set. The former is built by randomly choosing $C$ categories from the whole training set, with $K$ annotated samples of each, while the latter consists of another $n$ samples randomly selected from the same $C$ categories. This kind of few-shot task is usually designated as a ``$C$-way, $K$-shot'' task.\par
We present the support set as $S=\{(x_i,y_i)\}_{i=1}^{m=C{\times}K}$ and the query set as $Q=\{(x_j, y_j)\}_{j=1}^{n}$, respectively, where $x_{i/ j}$ indicates the image sample and $y_{i/j}\in\{1,\dots,C\}$ is the associated annotated category. It should be noted that the support set and query set share the same category label space, i.e. $\{1,\dots,C\}$, which is part of the whole database.\par
Similar to \cite{prototypicalnet,match_net,relation_net} the few-shot classifier we proposed consists of two modules: 1) feature embedding network $f_{\theta}$, which translates the image samples to the feature vectors, where $\theta$ represents the weights of the network; 2) feature distance function that evaluates the similarity of feature vectors. As in \cite{prototypicalnet}, we fix the feature distance function with squared Euclidean distance.\par
The training loss of each iteration is computed using the negative log-probability of the samples in query:
\begin{equation}
J(\theta)=-\sum_{j=1}^{n}log(p_\theta(y_j=\hat{y}_j|x_j))
\end{equation}
where $p_\theta(y_j=\hat{y}_j|x_j)$ indicates the probability of the sample $y_j$ belonging to its ground truth category $\hat{y}_j\in\{1,\dots,C\}$: 
\begin{equation}
p_\theta(y_j=\hat{y}_j|x_j)=\frac{exp(-d(f_{\theta}(x_j),c_{\hat{y}_j}))}{\sum_{\hat{y}_j^{'}}exp(-d(f_{\theta}(x_j),c_{\hat{y}_j^{'}}))}
\end{equation}
where $d$ is the function to calculate the squared Euclidean distance between the two inputs, $\hat{y}_j^{'}$ means all the categories involved in the few-shot task except $\hat{y}_j$, and $c_{\hat{y}_j}$ is the mean feature vector of samples belonging to the category $\hat{y}_j$ in the support set:
\begin{equation}
c_{\hat{y}_j}=\frac{1}{K}\sum_{y_i\in\hat{y}_j}f_\theta(x_i)
\end{equation}
\par
When testing, we compare the embedded feature vector of the under-test sample with all mean feature vectors of the candidate food categories included in the training set. The prediction $y_t$ is the category that exhibits the nearest squared Euclidean distance from the testing sample, which can be expressed as:  
\begin{equation}
y_t\gets{\mathop{\arg\min}_{y^h}}[d(f_\theta(x_t),c_{y^h})], y^h\in{C_h^m}
\end{equation}where $C_h^m$ is the set of candidate fine-grained categories belonging to the $m^{th}$ hyper category, and the value of $m$ is given by the prediction from the semantic segmentation described in the previous sub-section. Taking the hyper category of ``main course'' as an example, which corresponds to the case of $m=1$, the candidate set is therefore denoted as $C_h^1$ - that involves 139 fine-grained categories in total. To improve recognition performance,  $C_h^m$ can be further slimmed down by introducing the ``daily menu'' that only includes a few food categories ($\sim$7 per hyper category) served at a given date, with a compromise of updating the proposed system every day. 
\subsubsection{Network architecture}
\begin{figure}[t]
\centering
\includegraphics[width=1.0\linewidth]{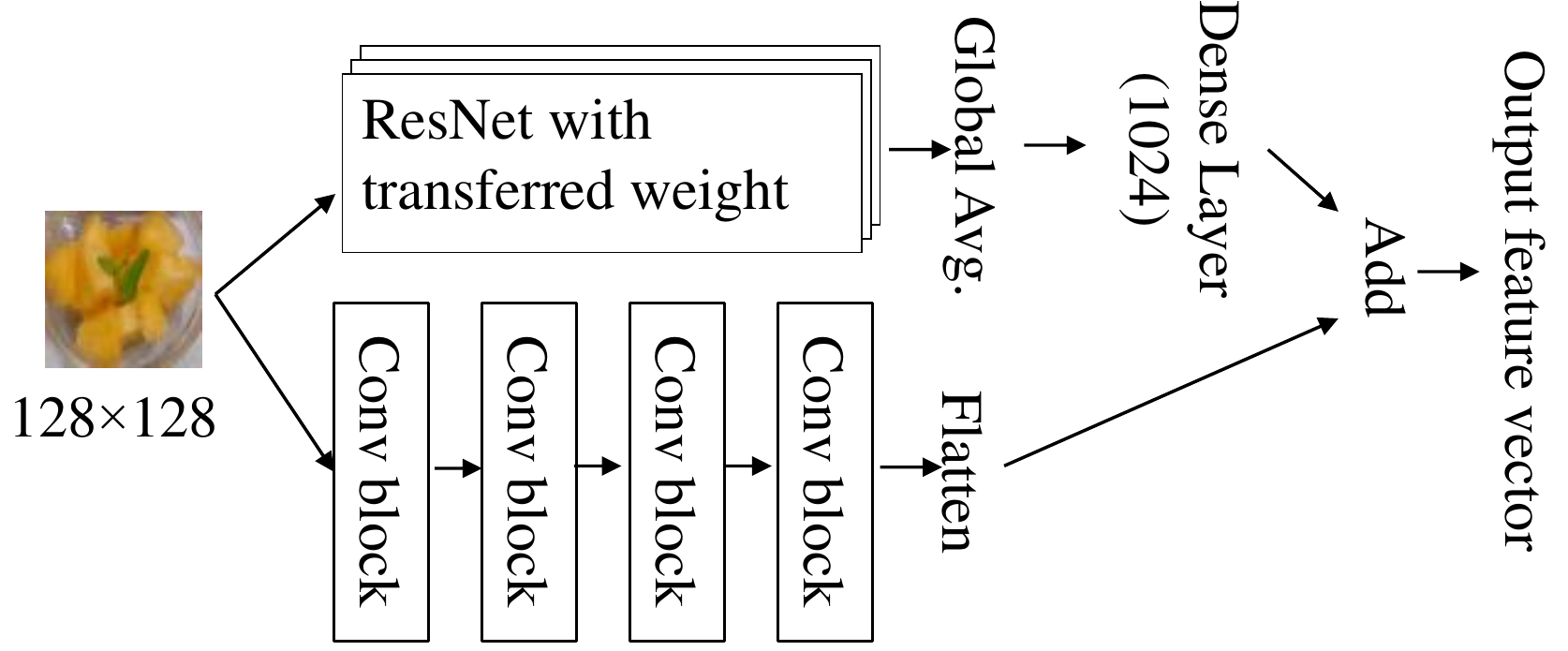}
   \caption{Parallel feature embedding network for few-shot learning. The transferred weights of the ResNet are fixed during network training.}
\label{fig:fewshot}
\end{figure}
In order to benefit from the sophisticated networks while avoiding over-fitting during training, a novel feature embedding network is proposed combining two parallel branches, as shown in Fig. \ref{fig:fewshot}. The upper branch applies the sophisticated ResNet50 architecture with transferred weights from ImageNet, which are fixed during training, while the lower branch utilises the widely used four convolutional blocks (4CONV) architecture \cite{prototypicalnet,match_net,relation_net} to prevent over-fitting. Each convolutional block contains a $64$-filter $3\times3$ convolution, batch normalisation, a RELU activation layer and a max-pooling layer. Note that the kernel size and the stride of the first max-pooling layer are both $4$, while that of the rest three layers are $2$. Finally, the output feature vector with length of $1024$ is obtained by adding the features extracted from both branches after a dense layer and a flatten layer, respectively.
\subsubsection{Consumed volume estimation}
The consumed volume of each food item is derived by simply subtracting the food volumes before and after the meal, which can be retrieved from the previously available food segmentation map and the depth map, associated with three stages: 1) 3D food surface extraction; 2) plate surface estimation; 3) volume calculation of each food item.\par
To calculate the volume of each food item, both the 3D food surface and the entire plate surface (which includes the area covered by food) are required. The depth image of the food area is firstly translated into a 3D point cloud and divided into triangular surfaces using the Delaunay triangulation method \cite{joachim3d}, which can be used to construct the 3D food surface. The plate surface estimation requires the previously built 3D plate model and the plate position, including location and orientation. The location of the plate is estimated using the plate segmentation map. The plate orientation is set as the normal vector of the tray plane, which is estimated using the RANSAC \cite{RANSAC} algorithm. Finally, the volume of each food item is calculated by simply integrating the height difference between the food surface and the plate surface along the normal vector of the food tray.\par
This method is sufficiently accurate to estimate the consumed volume, provided that the food is served with a normal plate (i.e. those shown in Fig. \ref{fig:plate}(a)-(d)). However, for packaged containers (shown in Fig. \ref{fig:plate}(e)), several additional heuristic methods are required. For instance, although it is difficult to identify the consumption of the sauce that is in the packaged containers (from Fig. \ref{fig:dbd} before and after meal), application of a heuristic rule can lead to a reasonable estimation. This might be that sauce consumption is proportional to salad consumption.\par
\subsubsection{Nutrient calculation}
To calculate the nutrient content of each food item, we firstly translate the food volume to the weight using (\ref{eq:vol}):
\begin{equation}
\label{eq:vol}
weight=\rho\cdot volume
\end{equation}
where $\rho$ is the density that is trained using linear regression for each fine-grained food category. Using the calculated weight associated with the nutrient recipes as provided by the central kitchen and the previously recognised food category, the nutrient content of each food item can be known for both before and after meal. The consumed nutrient content of each food item is simply calculated by subtracting the nutrient content after the meal from that before the meal. 

\section{Experimental results}
\label{sec:exp}

\subsection{Food segmentation}
In the experiments, the 5-fold cross validation evaluation strategy is adopted. For each fold, around 64 meals \footnote{The first 4 folds have 64 meals for testing, while the last fold has 66 testing meals.}  are used for testing, while the residual meals are split into training set and validation set with a ratio of 7:1. \par
The segmentation network is trained with the ``Adadelta'' optimiser and the categorical cross entropy loss. The loss weight of the ``plate'' and ``food'' segmentation branch are set as 0.6 and 1.0, respectively. The initial epoch number is set as 100, and the training process terminates when the validation loss stops decreasing for 10 epochs. The batch size is set as 4. To increase the image variability, we augment the input images by applying left-right and up-down flips during training.\par
Three kinds of metrics are utilised here to evaluate the semantic food segmentation performance, including: 1) the Intersection of Union (IoU) and mIoU, where IoU is evaluated for each hyper food category and mIoU is the average of all the individual IoUs, 2) the pixel level accuracy metric, and 3) the region-based F-score metric, which evaluates the worst ($F_{min}$) and the average ($F_{sum}$) segmentation performance \cite{joachim_seg, madima17}. Note that all the evaluations introduced above are only applied on the food areas in the images captured before the meal, and for all metrics a higher value indicates better performance..\par
\begin{figure}[t]
\centering
\includegraphics[width=1.0\linewidth]{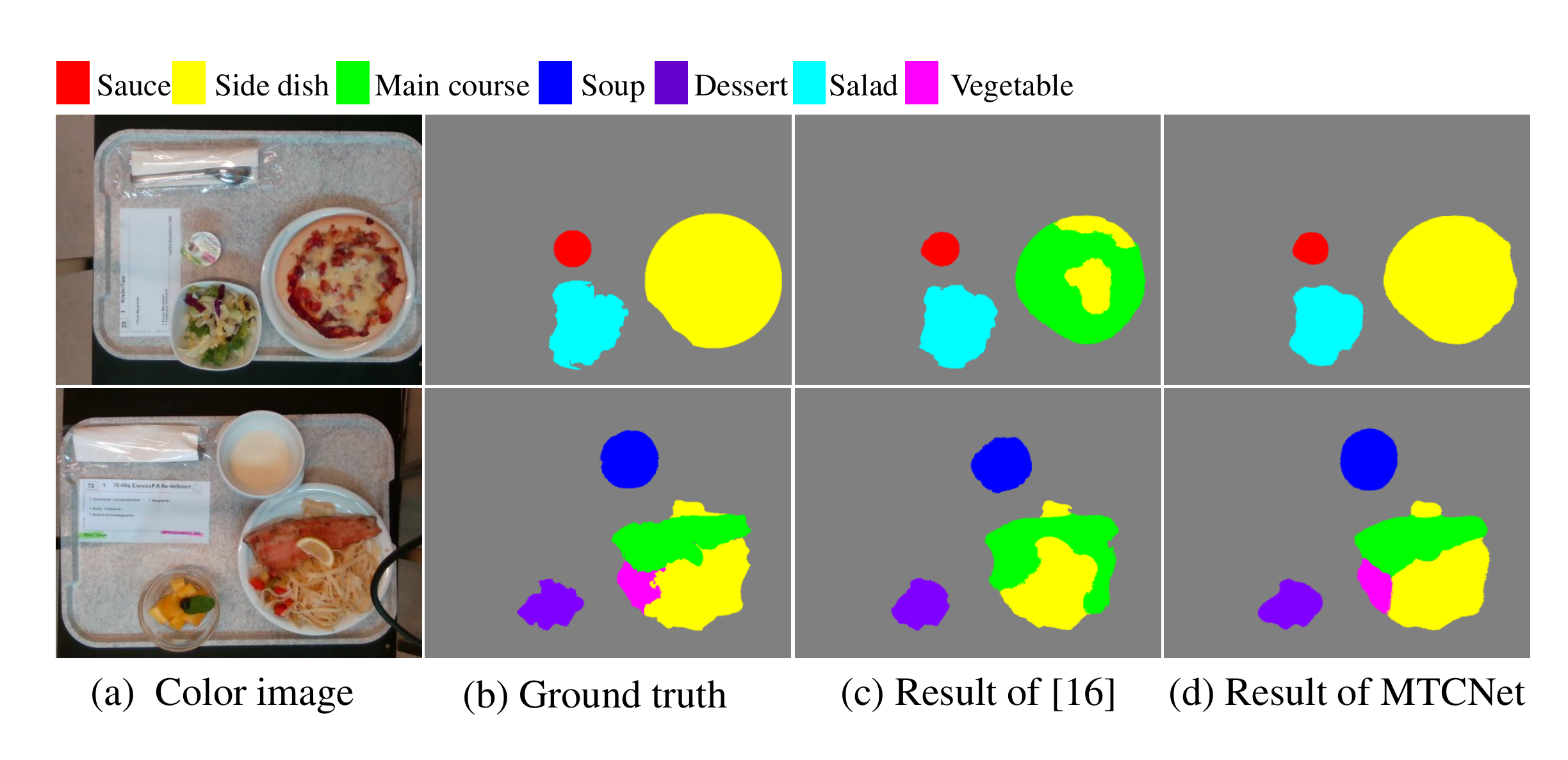}
   \caption{Food segmentation results using MTCNet, which outperform that obtained by  \cite{embc2019}.}
\label{fig:seg_r}
\end{figure}

The top part of Table \ref{table_seg} (first four rows) compares the metrics evaluated for the proposed MTCNet and our previous method \cite{embc2019} based on SegNet \cite{segnet}, with or without CRF post-processing \cite{dense_crf} for each. It can be found that the proposed MTCNet performs significantly better for all the metrics, especially after applying the CRF. This can be visually demonstrated as exemplified by two typical cases in Fig. \ref{fig:seg_r}, where the network in \cite{embc2019} wrongly predicts the meat as ``main dish'' (see the first row) and fails to distinguish the small vegetable area from the ``side dish'' (see the second row), since the receptive field is so small that the network overlooks the details in small regions. Nevertheless, both cases are correctly predicted using the proposed MTCNet, in which the pyramid feature encoding module provides a larger receptive field. \par
The second part of Table \ref{table_seg} (the last three rows) shows the evaluation results by applying the ablation study for the proposed MTCNet without CRF post-processing, in order to demonstrate the usefulness of different newly proposed components in our network. It can be seen that, after removing the CTLayer, all the metrics decrease with respect to those evaluated from the original MTCNet (the 3rd row in Table I). The performance further degrades after removing the ``plate branch'', validating the importance of the multi-task learning strategy in the network. More performance degradation can be found by removing the PreTrained weights in ResNet50, while training the network from scratch, demonstrating the necessity of the pretrained network when dealing with a small database \cite{transfer_learning}.
\begin{figure}[t]
\centering
\includegraphics[width=1.0\linewidth]{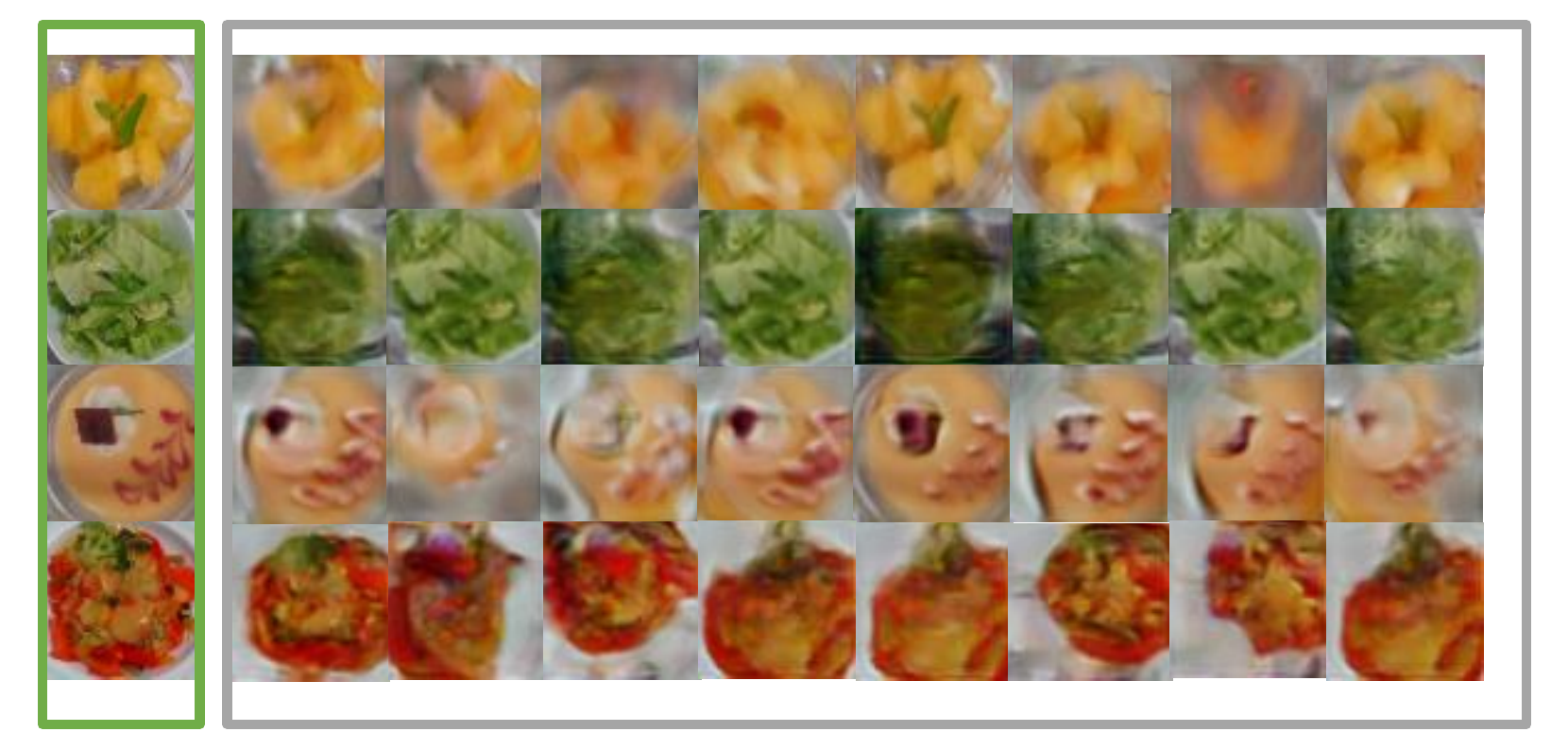}
   \caption{Image samples generated by GAN. The samples marked with green rectangle are real samples, while the samples in the gray box are the associated fake samples generated by GAN.}
\label{fig:gan}
\end{figure}

\begin{table*}[h]
\renewcommand{\arraystretch}{1.3}
\caption{SEGMENTATION RESULTS}
\label{table_seg}
\centering
\begin{tabular}{c p{1cm}<{\centering}  p{1cm}<{\centering} c  p{1cm}<{\centering}  p{1cm}<{\centering}  p{1cm}<{\centering}  p{1cm}<{\centering}  p{1cm}<{\centering}|p{1.2cm}<{\centering}|c c}
\hline
\hline

 \multirow{2}{*}{METHOD} & \multicolumn{8}{c|}{IOU(\%)}& Pix. Acc. (\%)&\multicolumn{2}{c}{F-score (\%)}\\
\cline{2-12}
                   & Main course&	Side dish&	Vegetable&	Salad&	Sauce&	Soup&	Dessert&	mIoU& &$F_{min}$ & $F_{sum}$\\
 \cline{2-9}
Network in \cite{embc2019}& 48.41	&40.74&	41.91&	80.70&	55.52&	75.33&	67.41&	58.53&	74.51&	61.43&	83.27\\
Network in \cite{embc2019}+CRF& 51.47&	43.30&	45.87&	85.14&	64.36&	84.29&	78.00&	64.63&	77.03&	66.59&	86.57\\
MTCNet&	61.00&	50.14&	60.83	&86.67	&69.03&	87.00&	78.98&	70.52	&82.15&	74.67&	88.76\\
MTCNet+CRF&	\textbf{61.46}&	\textbf{50.91}	&\textbf{63.51}&	\textbf{88.40}&	\textbf{77.76}&	\textbf{89.46}&	\textbf{84.75}	&\textbf{73.75}&	\textbf{83.50}	&\textbf{78.96}	&\textbf{89.88}\\
\hline
\hline
No CTLayer&	60.26&	47.67&	58.61&	86.19&	67.65&	85.18&	76.91&	68.92&	80.80	&74.05&	88.65\\
No plate branch	&59.76&	48.99&	56.94&	86.02&	67.16&	85.64&	76.09&	68.65&	80.96&	72.27&	87.60\\
No PreTrain&	53.58&	42.35&	51.17&	86.07&	62.83&	80.44&	77.92&	64.90&	76.39	&65.51&	86.37\\
\hline
\hline
\end{tabular}
\end{table*}

\begin{table*}[h]
\renewcommand{\arraystretch}{1.3}
\caption{FOOD ITEM RECOGNITION}
\label{table_recog}
\centering
\begin{tabular}{c p{2cm}<{\centering}  p{1cm}<{\centering} |  p{1cm}<{\centering}  p{1cm}<{\centering}  p{1cm}<{\centering}  p{1cm}<{\centering}  p{1cm}<{\centering} p{1cm}<{\centering} p{1cm}<{\centering}|c}
\hline
\hline

METHOD & Feature embedding&	Daily Menu&	Main course&	Side dish&	Vegetable&	Salad&	Sauce&	Soup&	Dessert	&Mean acc.\\
\hline
Prototypical Net &	4CONV &	N &	45.10 &	37.25 &	60.00 &	64.70 &	40.00 &	44.44 &	59.09 &	50.07 \\       
Proposed&	ParallelNet	&N&	\textbf{60.78}&	\textbf{56.86}&	\textbf{84.00}&	\textbf{82.35}&	\textbf{57.50}&\textbf{51.80}&	\textbf{72.27}&	\textbf{66.50}\\
\hline
Prototypical Net&	4CONV&	Y&	84.31&	80.39&	96.00&	94.11&	72.50&	85.19&	84.09&	85.22\\
Proposed&	ParallelNet&	Y&	\textbf{98.39}&	\textbf{92.15}&	\textbf{100.0}&	\textbf{100.0}&	\textbf{75.00}&	\textbf{88.89}&	\textbf{90.91}&	\textbf{92.19}\\
\hline
\hline
\end{tabular}
\end{table*}

\begin{table}[h]
\renewcommand{\arraystretch}{1.3}
\caption{Ablation study of GAN-based data augmentation}
\label{table_gan}
\centering
\begin{tabular}{c p{2cm}<{\centering}  c  p{1cm}<{\centering}}
\hline
\hline

METHOD & Feature embedding &	Augmentation &Mean acc.\\
\hline
Prototypical Net &	4CONV &	Standard&47.58 \\  
Prototypical Net &	4CONV &	GAN&50.07 \\ 
\hline
Proposed&	ParallelNet	& Standard&	63.90\\
Proposed&	ParallelNet	& GAN&	\textbf{66.50}\\
\hline
\hline
\end{tabular}
\end{table}


\begin{table}[h]
\renewcommand{\arraystretch}{1.3}
\caption{Comparison between standard image classification paradigm and meta-learning based approach}
\label{table_meta}
\centering
\begin{tabular}{p{3cm}<{\centering} p{2cm}<{\centering}  p{1cm}<{\centering}}
\hline
\hline

METHOD & Feature embedding &Mean acc.\\
\hline
Standard classifier &	4CONV &40.52 \\  
Prototypical Net &	4CONV & \textbf{47.58} \\ 
\hline
\hline
\end{tabular}
\end{table}
\begin{figure*}[htb]
\centering
\includegraphics[width=1.0\linewidth]{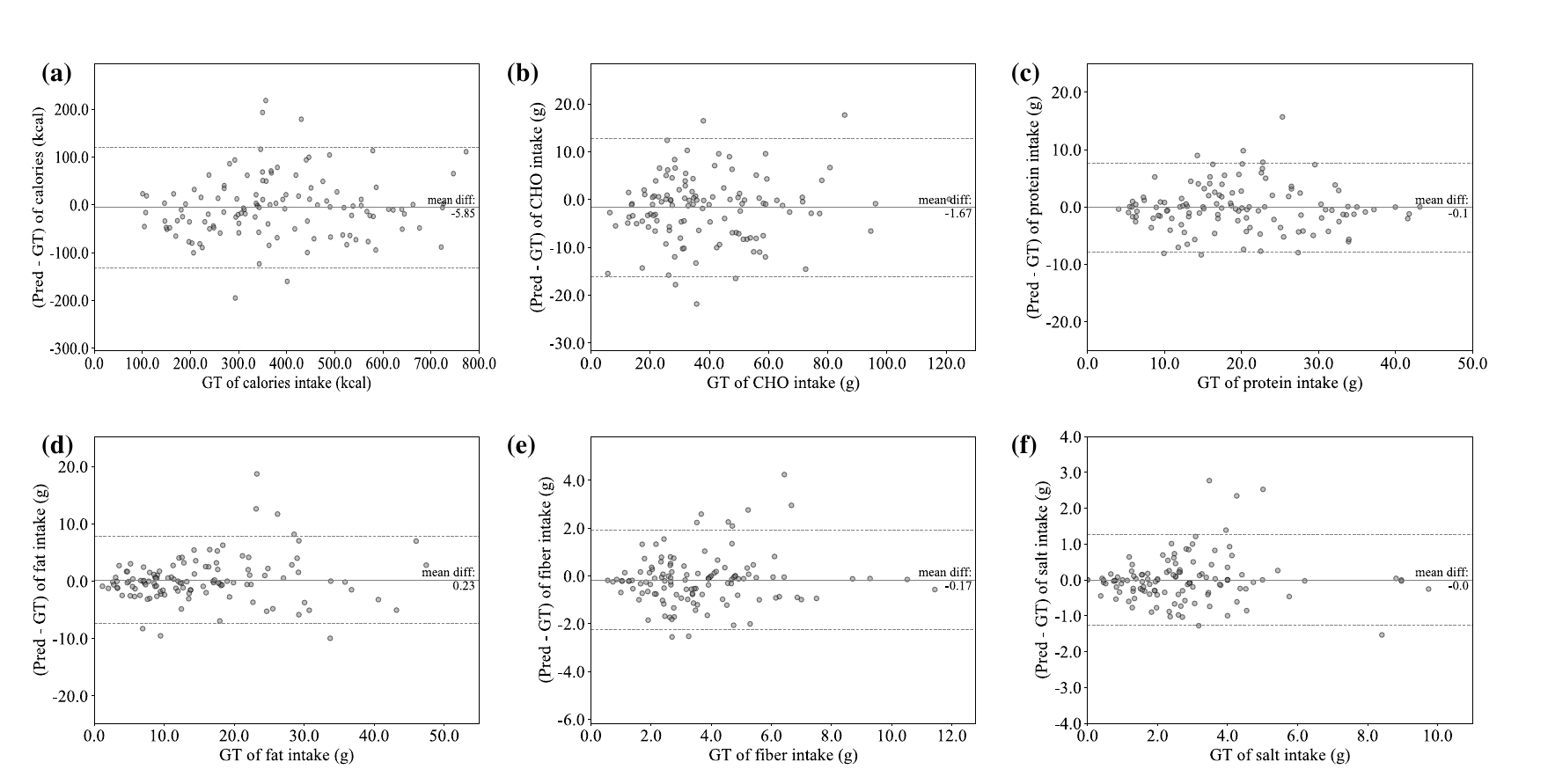}
   \caption{Bland-Altman plot of (a) calories; (b) CHO; (c) protein; (d) fat; (e) fibre and (f) salt intake. The horizontal axis of each plot indicates the ground truth value, while the vertical axis denotes the difference between the predictions of our algorithm and the ground truth. The middle lines indicate the mean difference (bias) between proposed algorithm and ground truth, while the dash lines mean the 95\% limits of agreement.}
\label{fig:ba}
\end{figure*}
\subsection{Fine grained recognition of food items}

For the food recognition experiments, 60 meals are selected for testing, and the remaining 262 meals are allocated for training. Here we ensure that all the food categories in the testing set have corresponding samples in the training set. In other words, food categories that only correspond to one sample in the database will not be chosen as testing meals. \par

The fine-grained food classifier is trained with the 10-way, 1-shot few-shot tasks, and there is 1 sample in the query set for each task. The training and testing images are resized into $128\times128$. During training, around 10\% food categories of the training set are split to a validation set. The network is trained using the ``Adam'' optimiser with learning rate of 1e-4 and weight decay of 1e-5. The batch size is set as 16 and the network is trained for 20$k$ iterations.\par
Data augmentation is one of the most commonly used strategies during the CNN model training \cite{review_ref1, review_ref2, alex_net}.  Standard data augmentation strategies, including flips, rotations, translations and Gaussian noise, can improve the generalization ability of the CNN model \cite{alex_net}. Recently, the Generative Adversarial Networks (GAN)-based approach has been investigated for data augmentation \cite{DAGAN,gan_aug}. The GAN-based approach gives more promising results than the standard approach.
In our experiments, the training data is augmented using a  GAN-based approach \cite{DAGAN}, which generates 8 fake samples for each original training sample, as exemplified in Fig. \ref{fig:gan}. \par

The performance of the proposed method is compared with the prototypical net \cite{prototypicalnet}, which is a standard few-shot learning baseline and has a similar architecture to the proposed method, except for the feature embedding network. The code of the prototypical net used in our experiments is from the official release of \cite{prototypicalnet}, and it is trained using the same setup with the proposed method. The usage of ``daily menu'' is quantitatively evaluated during the experiments.\par
Table \ref{table_recog} lists the evaluated recognition accuracies obtained by both the proposed method and the prototypical net, without and with the ``daily menu'' mentioned in Section \ref{sec:method}. As a benefit of the novel ParallelNet, the proposed method offers much greater accuracy with respect to the baseline. It is also found that the use of ``daily menu'' can dramatically boost the performance.\par

Table \ref{table_gan} presents the ablation study on the usage of GAN-based data augmentation. Comparison of the results indicates that the GAN-based data augmentation strategy outperforms the standard one by $\sim$ 3\% for both the prototypical net \cite{prototypicalnet} and the proposed network on our database. \par

To show the advantage of the meta-learning based framework with only a few training samples, the standard image classification  procedure and meta-learning based approach are compared (see Table \ref{table_meta}). The ``Standard classifier'' applies the same feature embedding network architecture as in the prototypical net \cite{prototypicalnet}, while an additional dense layer with the ``softmax'' activation is implemented for the food category prediction. From the results in Table \ref{table_meta}, the meta-learning based approach is $\sim$ 7\% more accurate than the standard classifier. This demonstrates that the meta-learning based approach performs well on our database.

\begin{table*}[!t]
\renewcommand{\arraystretch}{1.3}
\caption{NUTRIENT INTAKE ESTIMATION}
\label{table_nutrient}
\centering
\begin{threeparttable}
\begin{tabular}{c p{3cm}<{\centering}  p{2cm}<{\centering}   p{2cm}<{\centering} | p{3cm}<{\centering}  p{2cm}<{\centering}  p{2cm}<{\centering} }
\hline
\hline
&\multicolumn{3}{c|}{\textbf{With} daily menu} & \multicolumn{3}{c}{\textbf{Without} daily menu}\\
\cline{2-7}
&MAE(SD)&	MRE (\%)	&Correlation coefficient\tnote{**a}	&MAE(SD)&	MRE (\%)	&Correlation coefficient\tnote{**a}\\
\hline
Calories&	47.59 (43.55) kcal	&14.84&	0.923&	77.75(93.31) kcal	&18.52&	0.824\\
CHO&	5.33 (5.28) g	&18.16&	0.937&	6.96(7.75) g&	19.12&	0.914\\
Fat&	2.53 (2.89) g	&19.86&	0.927&	5.53(7.26) g&	27.70&	0.682\\
Protein&	2.87 (2.74) g&	17.00&	0.921&	3.67(4.13) g&	19.34	&0.889\\
Salt&	0.42 (0.49) g&	17.72&	0.941	&0.70(1.13) g	&20.20&	0.808\\
Fiber&	0.75 (0.76) g&	19.42&	0.910	&1.01(1.11) g&	21.99&	0.883\\
\hline
\hline
\end{tabular}
\begin{tablenotes}
\item[a] 0-0.19: very weak; 0.20-0.39: weak; 0.40-0.59: moderate; 0.60-0.79: strong; 0.80-1.0: very strong
\item[**] $p<0.001$
\end{tablenotes}
\end{threeparttable}
\end{table*}

\subsection{Assessment of nutrient intake }
In this section, we use the same training and testing dataset as that for the food recognition task, which involves 60 meals for testing and 262 meals for training. For each testing meal, the images captured before, during and after the meal are included, leading to two meal intake pairs which are the ``before-during'' pair and ``before-after'' pair. Thus, 120 meal pairs in total are used in this section for assessing nutrient intake.\par
To evaluate the nutrient intake estimation through the entire pipeline, the following steps are conducted: 1) the semantic segmentation network is retrained using the training set defined in this section, which is then applied to the testing images with CRF post-processing, 2) the corresponding food items in the two images for each meal pair are matched according to the information of the hyper food category output from the step 1, 3) the fine-grained categories are recognised using the proposed few-shot food classifier trained with GAN-based data augmentation and 4) the volume and nutrient intake of each meal pair are calculated using the method described in Section \ref{sec:method}.C and D. Note that the first three steps require processing the RGB images using machine learning algorithms, while the last step mainly focuses on processing the depth image.\par
Table \ref{table_nutrient} presents the evaluated mean absolute error (MAE), mean relative error (MRE) and correction coefficient \cite{pearson} of the predicted calories and 5 different types of nutrient intake, for the cases of ``with'' and ``without'' the daily menu. Whilst the case of without daily menu already shows good accuracies, the use of daily menu further boosts the performance - as ``very strong'' correlation ($r > 0.9, p < 0.001$) between the prediction and ground truth can be found for all the nutrient types, and the MREs are all lower than 20\%. Note that due to the absence of identical evaluation data, the strict quantitative comparison between the proposed method and other state of arts is not possible. However, according to the results reported by several similar studies (a CHO MRE of $\sim$25\% with the AI-based approach \cite{gocarb_clinical} and $r = 0.6489$ for intake calories estimation with traditional digital photography approach by professional health care staff \cite{study_6}), the proposed approach is expected to have better performance in terms of accuracy and efficiency.\par
To intuitively demonstrate the performance of the nutrient intake assessment, Bland-Altman plots \cite{bland_altman} for all the nutrient types are illustrated in Fig. \ref{fig:ba}(a)-(f), respectively. For each plot, the horizontal axis indicates the ground truth value of the nutrient intake, while the vertical axis denotes the errors, i.e., the difference between the prediction and the ground truth. The middle line in each plot shows the mean error (i.e. bias), while the two dashed lines represents the 95\% limits of agreement.\par
To figure out the impact of different processing steps on the overall pipeline error reported in Table \ref{table_nutrient}, two control variate experiments were conducted: 1) food segmentation ground truth was applied during evaluation, so that the pipeline error was attributed to the recognition and the volume estimation and 2) both food segmentation and recognition ground truths were applied during evaluation, so that the pipeline error was only attributed to the volume estimation. Both experiments were conducted with the daily menu information, and the evaluated results are reported in Table \ref{table_nu_sub1} and Table \ref{table_nu_sub2}, respectively. Both tables show very small differences from Table \ref{table_nutrient}, indicating the outstanding performance of the algorithms designed for both food segmentation and recognition, and in good agreement with the experimental results reported in Sections \ref{sec:exp} A and B. It can also be concluded that the volume estimation accounts for most of the errors in the whole pipeline (ca. 15\%), which may be because of the following two aspects: 1) the accuracy of the 3D surface construction algorithm is compromised in case of specular reflection, e.g. when there is strong light reflected from the liquid surface; 2) the food density inhomogeneity that imposes errors when converting the volume to the weight. Despite these intrinsic and practically inevitable errors, the proposed pipeline approach still offers unprecedented accuracy of the food nutrient intake estimation, thanks to the dedicated NIAD database and proposed AI algorithms.

\section{Conclusion}
In this paper, we have presented the design, development and evaluation of a novel AI-based automatic system for estimating nutrient intake for hospitalised patients in a pipeline manner. Several novel approaches are put forward, such as the new multimedia-nutrient combined database that collected data in the real hospital scenario, the dedicated designed MTCNet for food segmentation and the newly proposed few-shot learning classifier for food recognition. We demonstrated the prominent performance of the proposed algorithms comparing with the state-of-art in all the aspects, along with the feasibility of building an accurate nutrient intake assessment system with only small quantity training data.\par
Although the proposed software has been developed and evaluated using the images captured by the PC driven depth camera, it can easily be transferred to the smartphone - equipped with depth sensor for ultimate convenience.

\begin{table}[!t]
\renewcommand{\arraystretch}{1.3}
\caption{NUTRIENT INTAKE ESTIMATION WITH CORRECT SEGMENTATION$\ddagger$}
\label{table_nu_sub1}
\centering
\begin{threeparttable}
\begin{tabular}{c p{3cm}<{\centering}  c   p{2cm}<{\centering} }
\hline
\hline
&MAE(SD)&	MRE (\%)	& Correlation coefficient\tnote{**}\\
\hline
Calories&	46.05 (43.26) kcal	&14.64&	0.928\\
CHO	&5.33 (5.24) g&	18.09&	0.935\\
Fat&	2.30 (2.58) g&	17.93&	0.943\\
Protein&	2.69 (2.61) g&	16.16&	0.929\\
Salt&	0.42 (0.50) g&	17.27&	0.944\\
Fibre&	0.64 (0.57) g&	17.56&	0.939\\
\hline
\hline
\end{tabular}
\begin{tablenotes}
\item[$\ddagger$] with daily menu
\item[**] $p<0.001$
\end{tablenotes}
\end{threeparttable}
\end{table}

\begin{table}[!t]
\renewcommand{\arraystretch}{1.3}
\caption{NUTRIENT INTAKE ESTIMATION WITH CORRECT SEGMENTATION AND RECOGNITION$\ddagger$\tnote{$\ddagger$}}
\label{table_nu_sub2}
\centering
\begin{threeparttable}
\begin{tabular}{c p{3cm}<{\centering}  c   p{2cm}<{\centering} }
\hline
\hline
&MAE(SD)&	MRE (\%)	& Correlation coefficient\tnote{**}\\
\hline
Calories&	43.95 (42.24) kcal&	14.06&	0.931\\
CHO	&5.05 (5.09) g&	16.64&	0.942\\
Fat	&2.05 (2.53) g&	15.65&	0.948\\
Protein&	2.42 (2.44) g&	14.59&	0.941\\
Salt&	0.39 (0.49) g&	15.41&	0.950\\
Fiber&	0.52 (0.48) g&	14.23&	0.959\\
\hline
\hline
\end{tabular}
\begin{tablenotes}
\item[$\ddagger$] with daily menu
\item[**] $p<0.001$
\end{tablenotes}
\end{threeparttable}
\end{table}


%

\appendices
\begin{figure*}[htb]
\centering
\includegraphics[width=1.0\linewidth]{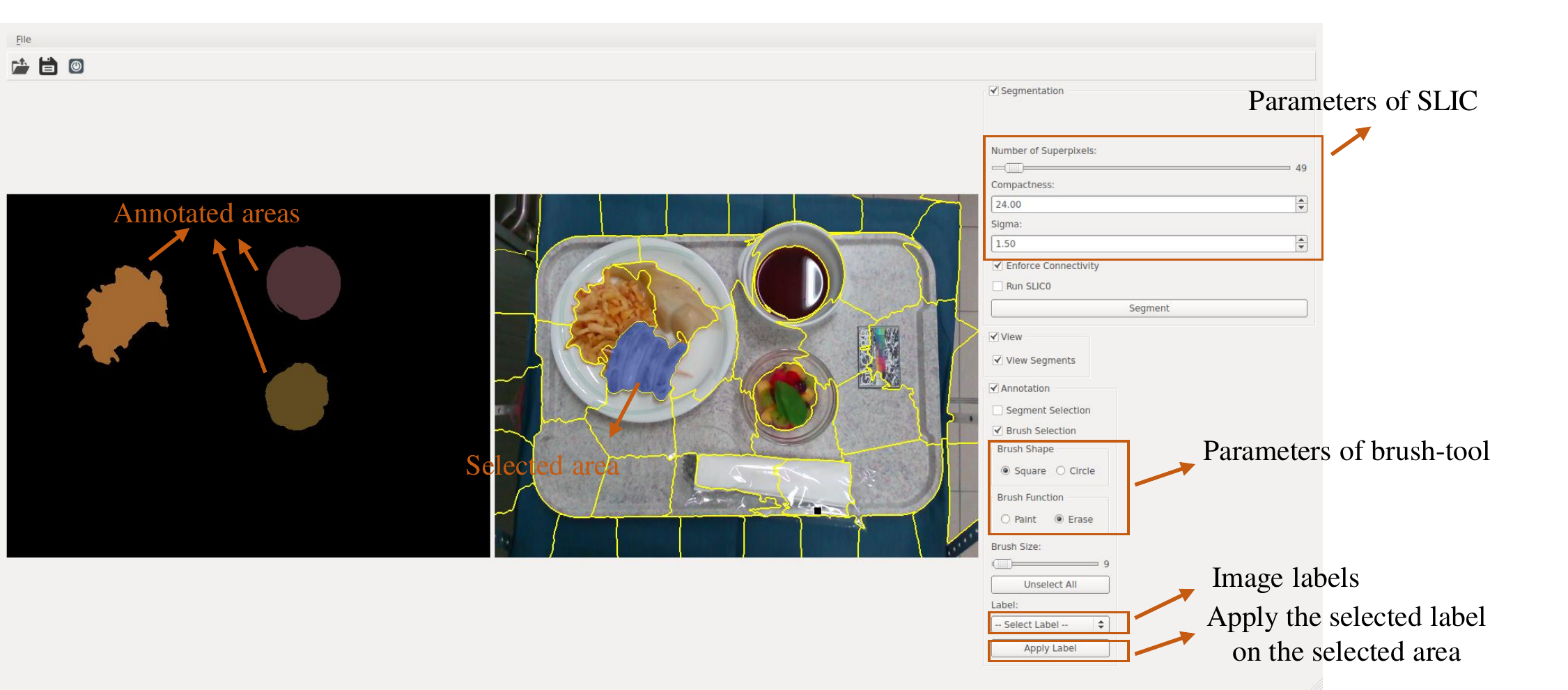}
  \caption{The interface of the in-lab developed segmentation tool.}
\label{fig:seg_tool}
\end{figure*}
\section{Annotation tool for food segmentation}
\label{app:tool}
The dataset has been annotated using an in-lab developed segmentation tool. The tool implements the SLIC algorithm \cite{data_anno} which generates superpixels based on the pixels{'} colour and spatial similarity, effectively color segmenting the image. Several parameters of the algorithm can be tuned using the user interface (GUI), such as the size and number of the segments, their compactness etc. (see Fig. \ref{fig:seg_tool}). Each segment can then be selected and annotated as one of those predefined by the user labels. Additionally the user can select or de-select pixels by using a brush-tool of varying size, which allows more detailed selection. The selected pixels can once again be labeled. Finally any selected area can at anytime be un-labeled, in case of a mistake.


\section*{Acknowledgment}

We would like to thank the Central Kitchen of the University Hospital ``Inspelspital'' and particularly Beat Blum, Thomas Walser, Vinzenz Meier for providing the menus, the recipes and the meals of the study. We would also like to thank all the volunteers who contributed in the image acquisition and annotation processes.

\ifCLASSOPTIONcaptionsoff
  \newpage
\fi



\bibliographystyle{IEEEtran}
\bibliography{IEEEabrv,egbib}
\end{document}